\pdfoutput=1

\documentclass[11pt]{article}

\usepackage{acl}
\usepackage{graphicx} 

\usepackage{amsmath}
\usepackage{times}
\usepackage{latexsym}
\usepackage{multirow}
\usepackage[T1]{fontenc}
\usepackage{hyperref}

\usepackage[utf8]{inputenc}
\usepackage{tabularx}
\usepackage{microtype}

\title{Text Difficulty Study: Do machines behave the same as humans regarding text difficulty?}

\author{Bowen Chen, Xiao Ding, Li Du, Qin Bing, Ting Liu \\
  Harbin Institute of Technology \\
  \texttt{\{xding, bwchen, ldu, qbing, tliu\}@ir.hit.edu.cn}}

\begin{document}
\maketitle
\begin{abstract}
Given a task, human learns from easy to hard, whereas the model learns randomly. Undeniably, difficulty insensitive learning leads to great success in NLP, but little attention has been paid to the effect of text difficulty in NLP. In this research, we propose the Human Learning Matching Index (HLM Index) to investigate the effect of text difficulty. Experiment results show: (1) LSTM has more human-like learning behavior than BERT. (2) UID-SuperLinear gives the best evaluation of text difficulty among four text difficulty criteria. (3) Among nine tasks, some tasks' performance is related to text difficulty, whereas some are not. (4) Model trained on easy data performs best in easy and medium data, whereas trains on a hard level only perform well on hard data. (5) Training the model from easy to hard leads to fast convergence.\footnote{Our code is available at \url{https://github.com/X3F4697J/ExamReadingBehavior}}
\end{abstract}

\section{Introduction}
Previous research in psycholinguistic shows \emph{language learner perform better on language tests when they start from easy sentences}(\citealp{textdiffculty1};\citealp{textdiffculty2};\citealp{textdiffculty3}) 
, which leads to a natural question that \emph{does machine behaves like a human concerning the text difficulty?} 
In this paper, we investigate how text difficulty affects models.  Specifically, we aim to answer the following questions:

(1)In which criteria of text difficulty, neural-based or feature-based or information-theory-based, the model behave like a human?

(2)In which model type, transformer\citep{vaswani2017attention}-based or RNN\citep{Rumelhart1986LearningRB}-based model, the model behave like a human?

(3)In which kind of task,  classification or regression task,  the model behave like a human?

\section{Text Difficulty}
(1) Flesch-Kincaid Score\citep{Kincaid1975DerivationON}: Given a document $d$, the number of sentences , word and syllables are $d_s$, $d_w$ and  $d_l$, then equation of Flesch-Kincaid Score is:
\vspace{-0.15cm}
\begin{equation}
\begin{split}
Flesch(d) = &206.835-\dfrac{1.015d_{w}}{d_s}-\dfrac{84.6d_{l}}{d_w}   
\end{split}
\end{equation}
\par
\vspace{-0.15cm}
\noindent $206.835$, $1.015$ and $84.6$ are empirical values.

(2)Neural Evaluation: Training model on datasets like Weebit\citep{inproceedings}, One-stop Corpus\citep{vajjala-lucic-2018-onestopenglish}, in which human experts judge the text difficulty. We fine-tune  BERT\citep{devlin2019bert} on the One-stop corpus to rank the text difficulty.

(3)Uniform Information Density (UID; \citealp{Fq};\citealp{10.3115/1073083.1073117};\citealp{doi:10.1177/00238309040470010201};\citealp{NIPS2006_c6a01432}) hypothesis: UID is based on the information theory\citep{6773024}, which the cognitive processing load of words is ratio to its log-probability and the ideal distribution of information should be uniform. A sentence that follows UID will not be cognitively taxing. We test two UID hypothesis operationalizations: UID Super-Linear(UID-SL) and UID Variance(UID-Var).

(1) UID Super-Linear:
\vspace{-0.3cm}
\begin{equation}
UID(\textbf{u})^{-1}=\dfrac{1}{N}\sum_{n=1}^ns(u_n)^k
\end{equation}
\par
\vspace{-0.3cm}
$k$ controls the strength of super-linearity. UID-SL suggests the text difficulty increases regarding the exponential sum of sentence surprisal.

(2) UID Language-Variance:
\vspace{-0.3cm}
\begin{equation}
UID(\textbf{u})^{-1}=\dfrac{1}{N}\sum_{n=1}^n(s(u_n)-\mu_{lang})^2
\end{equation}
\par
\vspace{-0.4cm}
$s(u_n)\overset{def}{=}-log\ p(u_n|u_{<n})$  means the  log-probability conditioned on its prior context in both equations.UID-Var suggests the text difficulty is decided by variance between the sentence surprisal and the mean language-level surprisal $u_{lang}$.We follow the implementation of  \citet{meister2021revisiting}, which the $k$ is $1.25$ and $u_{lang}$ is $3.8845$.
\section{Human Learning Matching Index}
We propose Human Learning Matching Index (\textbf{HLM Index}) to answer above questions, which has 3 sub-indexes $I_{task}$, $I_{model}$ and $I_{criteria}$.\footnote{For details about split of dataset please refer to Appendix.} 

We split each dataset into 3 text difficulty level computed by different criteria, which corresponds to easy, medium and hard level. Given tasks $T=\{t_1\dots t_j\}$, models $M=\{m_1\dots m_k\}$ and criteria $C =\{c_1\dots c_l\}$. Under task $t_o$ , criterion $c_p$ and model $m_q$, a model trained on each difficulty level has test performance on each difficulty level of test set $P(t_o, c_p, m_q)=\{p_{e}^{opq}, p_{m}^{opq}, p_{h}^{opq}\}$. Then, we define a logical score function $s$:
\vspace{-0.3cm}
\begin{equation}
    s(p_e,p_i,p_s )= 
    \begin{cases}
    0.75, & p_e\geq p_m\geq p_h,\\
    0.375, & p_e\geq p_h\geq p_m\\
    0, & p_m\geq p_h\geq p_e\\
    0, & p_m\geq p_e\geq p_h\\
    -0.375, & p_h\geq p_e\geq p_m\\
    -0.75, & p_h\geq p_m\geq p_e 
    \end{cases}
\end{equation}
\par
\vspace{-0.3cm}
Then we compute the $I_{model}$ index as following:
\vspace{-0.4cm}
\begin{equation}
\begin{split}
I_{model}(m_k) =\dfrac{1}{JL}\sum_{j=1}^{j} \sum_{l=1}^{l}(s(P(t_j,c_l,m_k))+\\
0.25sgn(s(P(t_j,c_l,m_k))f(STD(P(t_j,c_l,m_k))))
\end{split}
\end{equation}
\par
\vspace{-0.4cm}
$STD$ means standard deviation, $f$ is a Sigmoid function.Input of both $STD$ and $s$ is the performance triplet in easy, medium and hard difficulty level $p_e$, $p_m$ and $p_h$, and $sgn$ is the Sign function that is to decide the sign of STD.  The reason to include STD is to consider dispersion of performance. As $s$ only considers logical relation, the performance gap between different difficulty levels is ignored. However, the $f$ is to prevent $STD$ from dominating the HLM Index.By replacing $m_k$ to $t_j$ or $c_l$, we  have $I_{task}$ and $I_{criteria}$. The reason to choose $0.75$, $0.375$, and $0.25$ is to make the HLM index has a maximum and minimum value of $ \pm 1$, which is straightforward to illustrate the results. Changing these parameters only affects the maximum value and does not affect the results.
 

\section{Experiments}
\subsection{Datasets and Models}
To cover the spectrum of NLP tasks as much as possible, we select SST2, MRPC (Single/Pair-wise Sentence Binary Classification), QNLI, RTE(Multiple Classification), and STS-B(Regression) from GLUE Benchmark\citep{wang2019glue}.  We also include ROC Story (Multiple Choice) \citep{mostafazadeh-etal-2017-lsdsem},WIKITEXT2(Language Modelling) \citep{merity2016pointer},SQUAD 2.0(QA) \citep{rajpurkar2018know},CoNLL2003 NER(Tagging) \citep{tjong-kim-sang-de-meulder-2003-introduction}.\footnote{We use WT2, SQUAD, CoNLL2003, ROC to denote WIKITEXT2, SQUAD 2.0, CoNLL2003 NER and ROC Story.}

 Models in this paper are BERT and LSTM \citep{article}, which represent parallel and recurrent NLP models.
 

\subsection{Main Results}
\vspace{-0.2cm}
\begin{figure*}[!tb]
	\centering%
	\includegraphics[width=1.70\columnwidth]{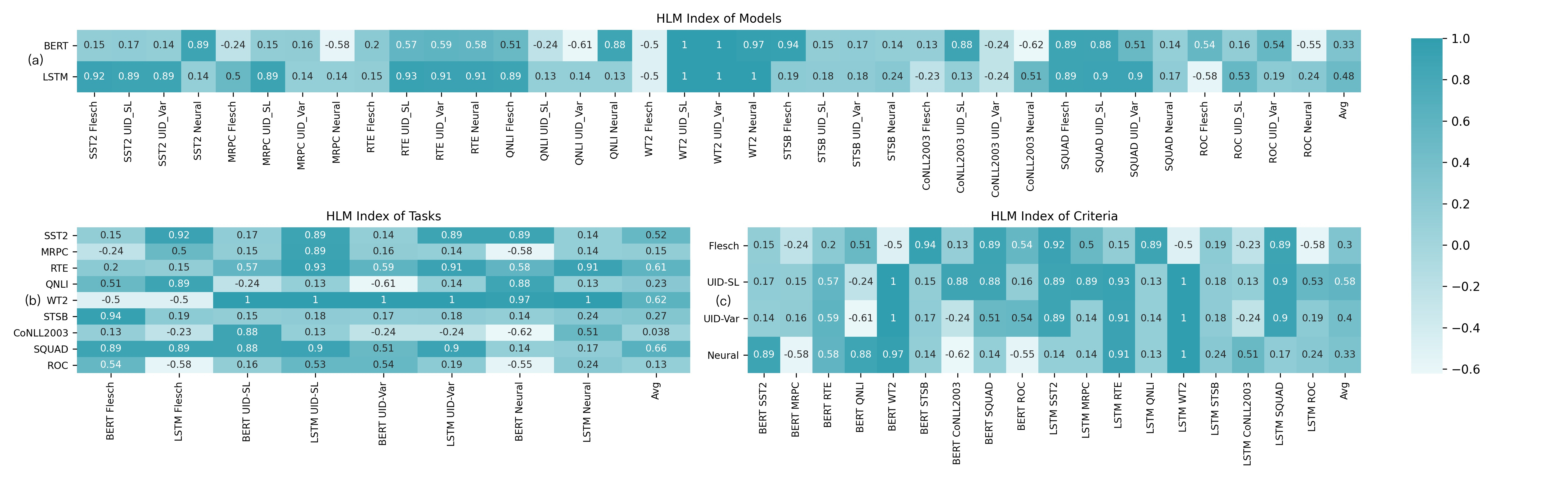}%
	\vspace{-0.4cm}
	\caption{HLM Index heatmap for models, criteria, and tasks. SL and Var represent super-linear and variance.}%
	\vspace{-0.6cm}
	\label{fig:whole}%
\end{figure*}%
\subsubsection{Models HLM Index}
 Figure \ref{fig:whole} (a) shows the $I_{m}$ of LSTM and BERT on different tasks divided by different criteria.

From the result, the LSTM has a higher average $I_{m}$ than BERT, which means the LSTM has a more human-like learning behavior. This contradicts \citet{merkx-frank-2021-human}, where they find the Transformer is more human-like with regard to learning human eye-tracking data, whereas we investigate it from learning behavior on multiple tasks. The results also contradict the idea that a model with higher task performance is more intelligent as the BERT model is known for the competitive performance with humans on several NLP tasks. The results suggest that a higher model performance does not guarantee that the model could also behave like a human. 
\subsubsection{Tasks HLM Index}
Figure \ref{fig:whole} (b) shows $I_{t}$ of different tasks on different splits. From the results, SQUAD and WT2 have the highest average, which means the performance of the LSTM and BERT on these tasks are highly related to the text difficulty even under different criteria. Especially in WT2, the $I_{t}$ reaches maximum $1$ in both LSTM and BERT in UID-SL, UID-Var and Neural criteria. However, the model gives a contrary score in the Flesch-Kincaid criterion, which means the sentence length and syllables are not a good criterion to split WT2 due to a strong tendency to give low difficulty to short sentences, which does not guarantee easiness. A high score in SQUAD indicates that the model's performance in QA is highly related to the difficulty of the context and question.

Moreover, CoNLL2003 and QNLI have a close zero score. For CoNLL2003, text difficulty is not an obstacle for NER as entities are mainly represented as uppercase in English, so the text difficulty does not influence the model or human to tag the entity. For QNLI, the model infers answers from context and question, which the difficulty is the inference rather than the text.

For other tasks, the $I_t$ are positive, which means the model performance on these tasks is also related to text difficulty but also intertwines with difficulty at a higher level like inference difficulty.
\subsubsection{Criteria HLM Index}
Figure \ref{fig:whole} (c) shows the $I_{c}$ of different criteria.

From the result, the UID-SL gives the highest match, whereas the Neural-based method gives the lowest match. The UID-SL gives the highest score, indicating that the text difficulty is better evaluated by the super-linear function, which is close to linear at the beginning and increases exponentially with the sentence surprisal.

Surprisingly, the Neural criteria yield the lowest match, and we expect the model to learn more sophisticated criteria from human experts as they could judge text difficulty from a higher level. This may be due to insufficient training data, and the model is not effectively learning human judgment from human experts and only learns surface features that cannot generalize well on unseen texts. Additionally, the UID-Var and Flesch-Kincaid give similar results, showing that these two criteria are relatively less expressive than UID-SL.
\vspace{-0.2cm}
\begin{table*}[h!t]\small

\centering
\begin{tabular}{c| c| c| c c c c c c c c c }\hline
Criterion&Model&Train& SST2 & MRPC & RTE & QNLI &WT2 & STS-B & CoNLL2003&SQUAD&ROC\\ \hline
\multirow{6}*{Flesch}&\multirow{3}*{BERT}&Hard&92.17& \textbf{84.24}&59.56&89.62&\textbf{68.83}& 83.61&89.92&81.53&85.27\\
&&Medium&\textbf{92.66}& 83.69&\textbf{63.18}&89.54&83.15& 85.69&\textbf{90.03} &81.57&84.55  \\
&&Easy&91.63& 84.01&61.15&\textbf{89.89}&89.44& \textbf{85.96}&89.89 &\textbf{82.05}&\textbf{86.08} \\
\cline{2-12}
&\multirow{3}*{LSTM}&Hard&84.69& 81.23&54.24&60.25&\textbf{156.24}&76.79&87.78&53.52&67.22  \\
&&Medium& 85.48& 81.22&\textbf{55.30}& 60.59&206.19&\textbf{78.31}&\textbf{87.91}&53.69&\textbf{66.90}  \\
&&Easy&  \textbf{86.32}& \textbf{81.37}&54.94&\textbf{60.77}&475.91&75.43 &87.48&\textbf{54.12}&65.36\\
\hline
\multirow{6}*{UID-SL}&\multirow{3}*{BERT}&Hard&91.17& 84.26&60.57&\textbf{89.96}&92.88&85.98&90.15&81.49&84.24 \\
&&Medium&\textbf{84.39}& \textbf{84.24}&59.35&89.62&86.36&\textbf{86.86}&90.21&81.66&\textbf{84.80}  \\
&&Easy&92.57&83.46&\textbf{62.38}&89.70&\textbf{67.90}&85.87&\textbf{90.32}&\textbf{81.69}&83.60 \\
\cline{2-12}
&\multirow{3}*{LSTM}&Hard&85.18&81.34&54.31&60.42& 522.19&73.86&87.48&53.20&66.96  \\
&&Medium&\textbf{85.66}&81.54& 55.65&\textbf{60.67}&199.31&\textbf{76.15}&88.42&53.60&65.98  \\
&&Easy&85.85&\textbf{82.02}&\textbf{56.48} &60.44&\textbf{166.33}&74.99&\textbf{88.79}&\textbf{54.36}&\textbf{67.27}\\
\hline
\multirow{6}*{UID-Var}&\multirow{3}*{BERT}&Hard&91.76&84.24&60.28&89.95&93.78&\textbf{87.72}&\textbf{89.98}&81.83&85.24  \\
&&Medium&\textbf{92.52}&\textbf{84.92}&58.92&\textbf{89.99}&86.19&86.95&89.75&81.55&84.76  \\
&&Easy&92.20&83.37& \textbf{63.39}&89.53&\textbf{74.84}&86.18&89.91&\textbf{81.95}&\textbf{86.24} \\
\cline{2-12}
&\multirow{3}*{LSTM}&Hard& 85.25&81.58&54.08&61.72&407.09&75.51&\textbf{88.21}&53.34&65.30  \\
&&Medium&85.66&\textbf{81.71}& 55.16& \textbf{62.06}&196.23&\textbf{76.39}&87.71&53.76&\textbf{67.82} \\
&&Easy&\textbf{85.85}&81.28&\textbf{55.67}&61.69& \textbf{166.41}&74.04&87.91&\textbf{54.43}&65.22\\
\hline
\multirow{6}*{Neural}&\multirow{3}*{BERT}&Hard&92.36&\textbf{86.54}& 61.22& 89.76&90.79&85.72&\textbf{90.32}&81.62&\textbf{87.90}  \\
&&Medium&92.50&85.62& 60.94& \textbf{89.84}&87.38& \textbf{85.75}&90.11&\textbf{82.10}&86.94  \\
&&Easy&\textbf{92.80}&84.74&\textbf{64.18}&86.53& \textbf{78.63}&85.12&90.07&81.81&84.77  \\
\cline{2-12}
&\multirow{3}*{LSTM}&Hard& 85.36&81.27&55.38&\textbf{60.94}& 304.66&75.40&87.62&53.14&69.18  \\
&&Medium&\textbf{85.82}&\textbf{81.73}&56.34&60.72&200.26&\textbf{78.13}&\textbf{88.79}&\textbf{54.98}&\textbf{70.88} \\
&&Easy&85.52&81.61&\textbf{56.91}&60.61&\textbf{198.69}&76.43&88.41&54.43&62.76\\
\hline
\end{tabular}
\caption{Results of models on different difficulties evaluated by four criteria. Accuracy for SST2, RTE, QNLI, and ROC. F1 for MRPC,SQUAD and CoNLL2003. Perplexity for WT2. Pearson correlation for STS-B. The HLM Index is computed based on this table.}
\vspace{-0.4cm}
\label{glue_results}
\end{table*}

\subsection{Effect of Training Order}
In this part, we train the model in a human-like schedule that trains from easy to hard and another schedule that trains reversely, then report the convergence step divided by the total steps and the performance on the test set. We select RTE, SQUAD, MRPC, and SST2 to perform the experiments. The results are in Table \ref{tab:train order}.\footnote{Default criterion to split the data is UID-SL.}
\begin{table*}[!tb]\small
    \centering
    \begin{tabular}{l|c|c|c|c|c|c|c|c|c|c|c|c}\hline
    \multirow{3}*{Task}&\multicolumn{6}{c|}{LSTM}&\multicolumn{6}{c}{BERT} \\
    \cline{2-13} 
    &\multicolumn{2}{c|}{E $\rightarrow$ H}&\multicolumn{2}{c|}{H $\rightarrow$ E}&\multicolumn{2}{c|}{Rand}
    &\multicolumn{2}{c|}{E $\rightarrow$ H}&\multicolumn{2}{c|}{H $\rightarrow$ E}&\multicolumn{2}{c}{Rand} \\
    \cline{2-13}
    &P&C(\%)&P&C(\%)&P&C(\%)&P&C(\%)&P&C(\%)&P&C(\%)  \\
    \hline
        SST2&$81.65$&$\textbf{15.07}$&$81.67$&$36.29$&$\textbf{81.77}$&$18.50$&$\textbf{93.00}$&$\textbf{22.49}$&$92.88$&$44.23$&$92.78$&$28.44$\\
    \hline
         MRPC&$82.06$&$\textbf{20.04}$&$81.91$&$53.22$&$\textbf{82.26}$&$40.16$&$\textbf{87.73}$&$\textbf{19.04}$&$87.43$&$43.24$&$86.59$&$31.09$\\
    \hline
         RTE&$\textbf{58.18}$&$\textbf{16.75}$&$58.19$&$75.38$&$58.12$&$42.33$&$64.62$&$\textbf{31.22}$&$63.89$&$54.23$&$\textbf{65.34}$&$44.17$\\
    \hline SQUAD&$56.32$&$\textbf{39.22}$&$56.21$&$68.63$&$\textbf{56.43}$&$47.21$&$82.34$&$\textbf{30.21}$&$82.03$&$53.21$&$\textbf{83.21}$&$43.46$\\
    \hline
    \end{tabular}
    \vspace{-0.2cm}
    \caption{Performance and convergence of different tasks. Rand means a randomly shuffled train set. P means performance. C means the convergent step divided by total steps, which is the lower, the better.}
    \vspace{-0.4cm}
    \label{tab:train order}
\end{table*}
\begin{figure}[!tb]
	\centering%
	\includegraphics[width=0.72\columnwidth]{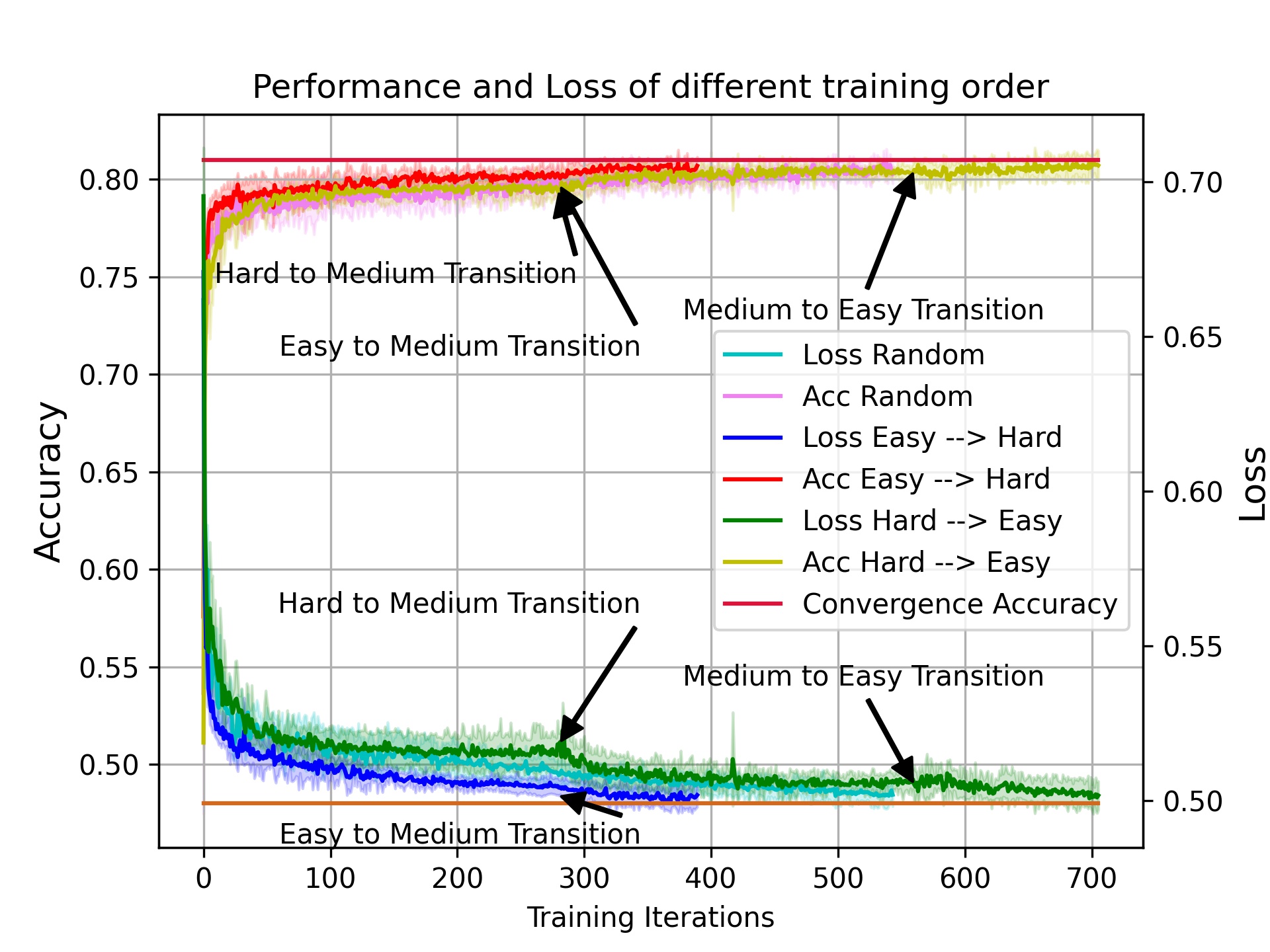}%
	\vspace{-0.4cm}
	\caption{LSTM learning curve in SST2 task. Transition points are annotated using arrow symbols.}%
	\vspace{-0.4cm}
	\label{fig:task case}%
\end{figure}%

From the results, training the model in a human-like schedule leads to a quicker convergence, whereas a reverse schedule shows the slowest convergence. The random schedule is in between. This means we could train the model more efficiently in a human-like schedule. Additionally, different training schedules give a close best performance, indicating the final performance is not sensitive to the training schedule, but the convergence is sensitive to it, which helps explain why a difficulty insensitive schedule is successful in NLP. However, even the best performance is insensitive to difficulty, but we could still reach comparable performance with fewer data in a  human-like training schedule. 
A case on the SST2 task is in Figure \ref{fig:task case}. We could see a human-like training order leads to an evident quicker convergence, whereas the reverse schedule leads to the slowest convergence. The LSTM uses the easy part and a subset of the medium part of train data in the human-like training schedule to achieve the best performance. In contrast, the reverse schedule needs to use whole data to achieve the same performance, implying that the model did not effectively learn the hard data.

\subsection{Transfer Between Text Difficulty}
This part investigates the difficulty transfer on the test set. For example, suppose a model trained on easy data performs better. In that case, we ask \emph{in which difficulty level of the test set, the performance improves or decreases ?} We split the test set of SST2, WT2, MRPC, RTE, QNLI, and SQUAD into 3 difficulty levels using all criteria then collects the model performances on each difficulty level of the test set. We sort the results and give $3,2$ and $1$ scores to the best, mediocre and lowest performances. The average score is in Table \ref{tab:transfer_table}.
\begin{table}[!ht]\small
    \vspace{-0.3cm}
    \centering
    \begin{tabular}{l|c|c|c|c|c|c}
    \hline
     \multirow{3}*{Train }&\multicolumn{3}{c|}{LSTM}&\multicolumn{3}{c}{BERT}\\
     \cline{2-7}
     &\multicolumn{3}{c|}{Evaluation Set}&\multicolumn{3}{c}{Evaluation Set}\\
     \cline{2-7}
     Set&Easy&Med&Hard&Easy&Med&Hard\\
    \hline
    Easy&\textbf{2.54}&\textbf{2.33}&1.29&\textbf{2.67}&\textbf{2.37}&1.67\\
     \cline{2-7}
     Med&2.16&2.21&2.17&1.95&2.24&1.83\\
     \cline{2-7}
     Hard&1.29&1.46&\textbf{2.54}&1.38&1.46&\textbf{2.50}\\
     \hline
    \end{tabular}
    \vspace{-0.2cm}
    \caption{Transfer scores between different text difficulties. Med means medium difficulty level.}
    \vspace{-0.4cm}
    \label{tab:transfer_table}
\end{table}

From the results, the model trained on the easy level has the best performance in both easy and medium levels, whereas trained on the hard level only performs well on the hard level and fails in other levels. Additionally, the model trained on a medium level gives a stable performance in all difficulty levels of the test set. The reason may be hard text difficulty dataset follows a different distribution from the easy and medium levels. In the hard dataset, as pointed out by results in Sec 4.2.2, the difficulty not only relates to the text level but also relates to a higher concept like the difficulty of inference or understanding, and a hard text is a feature of such difficulty. This suggests the distribution of the hard level might be different from easy and medium level, so the model trained on hard data does not perform well on other levels.

\section{Conclusion}
In this work, we investigated how and in what way the text difficulty affects and exists in NLP tasks and models. We analyzed experiments on nine tasks using HLM Index. Results show that LSTM gives more human-like behavior. UID-SL gives the best text difficulty evaluation. Some tasks are related to text difficulty, whereas some are not. Moreover, transfer experiment shows that the training begins with the easy data leads to a more general and better performance than hard data. Additionally, training with a human-like  schedule is more efficient than other schedule and leads to a quicker convergence.
\bibliography{anthology,custom}
\bibliographystyle{acl_natbib}
\end{document}